# Identification and Counting White Blood Cells and Red Blood Cells using Image Processing Case Study of Leukemia

**Esti Suryani, Wiharto, and Nizomjon Polvonov**

*Ilmu Rekayasa dan Komputasi Research Group, Sebelas Maret University*
*Jl. Ir. Sutami No 36 A Kentingan Surakarta*
suryapalapa@yahoo.com

*Ilmu Rekayasa dan Komputasi Research Group, Sebelas Maret University*
*Jl. Ir. Sutami No 36 A Kentingan Surakarta*
wi_harto@yahoo.com

*Ilmu Rekayasa dan Komputasi Research Group ,Sebelas Maret University*
*Jl. Ir. Sutami No 36 A Kentingan Surakarta*
kabutar571@gmail.com

**Abstract**

Leukemia is diagnosed with complete blood counts which is by calculating all blood cells and compare the number of white blood cells (White Blood Cells / WBC) and red blood cells (Red Blood Cells / RBC). Information obtained from a complete blood count, has become a cornerstone in the hematology laboratory for diagnostic purposes and monitoring of hematological disorders. However, the traditional procedure for counting blood cells manually requires effort and a long time, therefore this method is one of the most expensive routine tests in laboratory hematology clinic. Solution for such kind of time consuming task and necessity of data tracability can be found in image processing techniques based on blood cell morphology . This study aims to identify Acute Lymphocytic Leukemia (ALL) and Acute Myeloid Leukemia type M3 (AML M3) using Fuzzy Rule Based System based on morphology of white blood cells. Characteristic parameters witch extractedare WBC Area, Nucleus and Granule Ratio of white blood cells. Image processing algorithms such as thresholding, Canny edge detection and color identification filters are used.Then for identification of ALL, AML M3 and Healthy cells used Fuzzy Rule Based System with Sugeno method. In the testing process used 104 images out of which 29 ALL - Positive, 50 AML M3 - Positive and 25 Healthy cells. Test results showed 83.65 % accuracy .

**Keywords: ALL, AML, Fuzzy Rule-Based System, Granule Ratio, Nucleus Ratio, WBC Area, White Blood Cell Morphology.**

.

## I. Introduction

Biomedical research field in Indonesia is still relatively small, especially in fields related to clinical laboratories (Usman, K, 2008). Currently the ease, convenience and accuracy is something that is considered as a necessity. The development of information and communication technology, on the other hand, is opening new breakthroughs in various fields, including in the medical field. Several fields related to clinical laboratories are interesting and challenging topics of which are the blood cell image analysis. The analysis includes the computation effort, separation of various blood cells, through the analysis of the forms of blood cells to determine abnormalities that may occur. Blood cell abnormalities which are leukemia or blood cancer.

Leukemia is a cancer that effects (damages) the blood and bone marrow where blood cells are made (Leukaemia Foundation, 2011). Acute leukemia occurs when white blood cells proliferate abnormally rapidly and overflow into the blood stream (Leukaemia Foundation, 2011). According to the WHO ( Word Health Organization ) on December 2009 , Fact Sheet 4.1 with the title " Incidence of Childhood Leukemia " , Leukemia is a malignant disease that usually affects children , accounting for 30% of all cancers diagnosed in children under 15 years that live in Industrialised Countries. Similarly Indonesia is not an exception,as in the Asia-Pacific Journal of Cancer Prevention Vol. 12 , in 2011 , results of the World Health Organization research shows that the majority of patients diagnosed as leukemia positive is between the ages of 2-4 years as many as 497 of the 541 patients that suspected suffering from leukemia, (Supriadi, dkk, 2011) . There are four types of Leukemia, Acute Myeloid Leukemia (AML), Acute Lymphoblastic Leukemia (ALL), Chronic Myeloid Leukemia (CML), Chronic Lymphocytic Leukemia (CLL). CLL and AML generally occurs in adults while ALL generally occurs in children and grows rapidly ( Leukaemia Foundation, 2011)

The cause of leukemia is not known for sure. Leukemia can be diagnosed with complete blood cell count and compare the number of RBC and WBC. It is a cornerstone in the hematology laboratory and used to perform screening, case studies, and monitoring. However, the traditional procedure for counting blood cells with a microscope manually requires a lot of time and energy, and is one of the most expensive routine tests in clinical hematology laboratory ( Houwen, B, 2001). The best way to quickly and accurately calculate blood cells were suggested by WHO is using Immunophenotyping which WHO has been using for research in Indonesia for diagnosing leukemia. According to the WHO Immunophenotyping proven fit for detection of childhood leukemia (Supriadi, dkk, 2011).The way Immunophenotyping works is blood sample passed through the detector (flow-cytometer) and then discarded as waste where as in medical diagnosis itis very important to have tracking/recording of data forspeed of diagnosis.

Time consuming process and lack of trackable data problem can be overcome by using image processing techniques for counting blood cells from the pictures that have been taken with a microscope . Many studies conducted on the blood cell count with image processing techniques from which are research (Dorini, L.B, Minetto, R, and Leite, N.J, 2007) divides the white blood cells with analysis based on morphological concepts which separates the cell nucleus (nuclei) and cytoplasm. While in other studies (Habibzadeh, M, et al.2011), Immersion Watershed algorithm is proposed to calculate the red and white blood cells separately. Otsu and Niblack methods are used for image binarization. Then white blood cells are separated from the red blood cells (Red Blood Cell / RBC) based on RBC size estimates using granulometry. Devesh D. Nawgaje and Dr. Rajendra D.Kanphadeused Fuzzy Inference System (FIS) for detecting WBC in the study (Devesh D, Nawgaje, And Kanphade, R. D, 2011). FIS approach is used for edge detection in microscopic images of the bone marrow. Mamdani method is chosen as the defuzzification procedure. For the implementation of the system 8 - bit images are used as input and output for defuzzification. This way they could detect all leukocytes in the sample image. FIS algorithm successfully detected edges of the WBC. Research to Identify ALL based on the morphology of white blood cells has also been carried out by Fuzzy Rule Based Systemusing Sugeno method and obtained 73.68 % accuracyfrom testing 57 images which are 35 ALL -

Positive and 22 ALL – Negative. Characteristic parameter used to identify ALL are WBC Area , Nucleus Ratio , and Total count of granules (Suryani, E, Wiharto, and Polvonov, N, 2013).

This research proposes the idea of identification of Acute Lymphoblastic Leukemia and Acute Myeloid Leukemia using image processing techniques. Image segmentation using Canny Edge Detection algorithm, and improve it with Ellipse Detection method, Fuzzy Rule Based is used to identify chatactetistics of ALL, AML M3 or Healthy blood cells from captured digital bllood images.

## II.    Methods

Experiments in this study conducted through the stages shown in Figure 1 below:

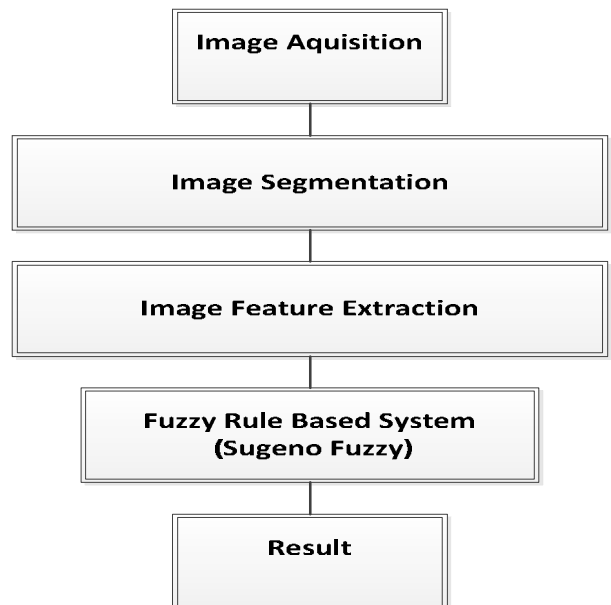

**Figure 1** Research Steps

### A. Image Acquisition

Digital Image of ALL and AML M3 in this case is obtained from a blood smear samples from 'Karang Anyar General Hospital', using a digital microscope with a magnification of 1000x in Biological Laboratory in Faculty of Mathematics and Natural Sciences, Sebelas Maret University. Tests conducted with AML M3 positive 50 image samples, 10 digital images samples of known as ALL positive, obtained from the capture of digital microscope with magnification of 1000x and has been confirmed by Doctors Pathologist Dr. Niniek Yusida, Sp.PK., M.Sc. also 19 ALL positive images and 25 ALL &AML  negative Images, obtained in the form of digital image JPG format with 24 bit color depth, resolution of 2592 × 1944 pixels which is captured by Power-Shot G5camera with magnification of 300x and 500x. Digital image data obtained from: Dr. Fabio Scotti, Università Degli Studi di Milano, Department of Information Technologies via Bramante 65, 26013 Crema (CR), Italy and has been confirmed by oncology experts.

### B. Image Segmentation

Image Segmentation stage aims to separate and detect white blood cells (WBC) and red blood cell (RBC). The first stage of image segmentation is todetectWBC. WBC detection stage is the most important part in this study, because the WBC will be used as a morphological characteristic to detect ALL and AML M3.

Morphological characteristics to be searched are: WBC area - size of the area that is the number of pixels nucleus and cytoplasm, Nucleus Ratio - the ratio of nucleus pixels and WBC area, and the last one is Granule Ratio is the ratio of granule pixels with pixels of nucleus.

### a. Color Filter

Color filters are used to extract WBC regions. Color main filter that will be used is purple color ("Giemsa"). The purple color ('Giemsa') is used in the 'blood smears' before usage (observing it) with a microscope.There are also two more color filters: 'dark blue' color filter used to extract WBC nucleus and 'reddish purple' color filter is used to extract granule especially found in AML M3 (Auer rods).

### b. Grayscale

After getting the WBC region, further Grayscale filters need to be used to reduce the color of digital image into 8 bits. This method is used to convert all colors to grayscale (gray) which will provide higher accuracy for the threshold.

### c. Thresholding

Thresholding phase is used to flatten the gray image on the WBC region that is to split between the background and the object in the image using Equation 1. Threshold value taken in this process is 1. The threshold function $f_{threshold}(a)$ maps all pixels to one of two fixed intensity value $a_0$ or $a_1$; i.e. (Burger, W, and Burge, M. J, 2009).

$$f_{threshold}(a) = \begin{cases} a_0 \ for \quad a, < a_{th} \\ a_1 \ for \quad a \geq a_{th} \end{cases}$$

$$\textbf{with } 0 < a_{th} \leq a_{max}$$

(1)

### d. Canny Edge Detection

Canny Edge Detection is used to detect edges that will result very thin and accurate edges. 'Canny Edge Detection' is known as edge detection algorithm that is the most accuratealgorithmand resulting edges are very delicate and thin. This process takes 3 inputs: 'Low Threshold', 'High Threshold', 'Sigma'. All three inputsare used to obtain maximum accurate results. The algorithm runs in 5 separate steps (Moeslund, T. B, 2009):

1. Smoothing: Blurring of the image to remove noise

2. Finding gradients: The edges should be marked where the gradients of the image has large magnitudes.

3. Non-maximum suppression: Only local maxima should be marked as edges.

4. Double thresholding: Potential edges are determined by thresholding.

5. Edge tracking by hysteresis: Final edges are determined by suppressing all edges that are not connected to a very certain (strong) edges.

*e. Circle Detection*

Circle Detection is used to detect circles in an image using the "inner and outer circle" method. From the edges of WBC itshigh determined and described two circles, the inner circle and the outer circle with a diameter of specified tolerance. If the edge of the cell is detected as a circle it will be counted as one cell, if not then it will be filtered by ellipse detection to detect cells that overlap. Equation 2 is the formula of "inner and outer circle" with a tolerance of 30 that will be used in this study:

$$Circle.Diameter_{inner} = WBC_{height} - \left(\frac{WBC_{height}}{100}\right) * \mathbf{30};$$

$$Circle.Diameter_{outer} = WBC_{height} + \left(\frac{WBC_{height}}{100}\right) * \mathbf{30} \qquad (2)$$

If WBC edges are between the two circles (red circle) and do not cross red circles then identified as a circle. Figure 2 is a picture example of "inner and outer circle"algorithm:

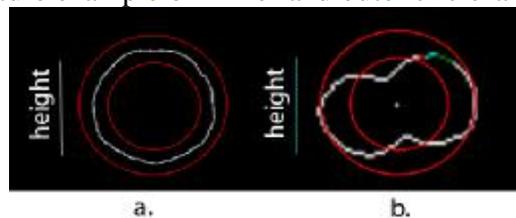

**Figure 2.** Circle Detection (a) WBC edges that are detected as a circle, (b) WBC edges that are not detected as a circle.

*f. Ellipse Detection*

WBC edges that are not detected as a circle will be filtered by 'Ellipse Detection' algorithm to detect cells detecting overlapping circles where each circle is counted as one cell. 'Ellipse Deteksion' algorithm detects curve segments in the edge image and choose any pair to test whether the curve segments related to the same ellipse or not and connect it as an ellipse if yes. In this paper we are going to use the new method (Hahn, K, et al. 2008), to detect the ellipse.

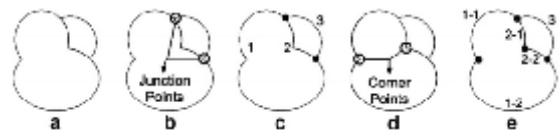

**Figure 3**. (a) occluded ellipse contour, (b) junction points in (a), (c) 3 curves that is some imperfect curve segments, (d) corner points to complete curve segments, (e) success to construct 5 curve segments.

First, construct three curves shown in Figure 3c using two junction points as in Figure 3b. However, curves 1 and 2 are each included in different elliptical contours. One needs to separate them as Figure 3e using two corner points in Figure 3d. Then the five curve segments are constructed perfectly..Having obtained the curve respectivelywill be inserted into the following merging measure formula to find the curves in the same ellipse:

$$MM = \begin{bmatrix} \ddots & & \\ & MM_{ij} & \\ & & \ddots \end{bmatrix}_{N \times N}$$

$$MM_{ij} = D(CS_i, CS_j)\Theta(CS_i, CS_j) \tag{3}$$

$MM$ is an $N$ by $N$ square symmetric matrix and $N$ is the number of curve segments. $MM_{ij}$ is merging measurement value between the $i_{th}$ curve segment and the $j_{th}$ curve segment (Equation 3). The return $D(CS_i, CS_j)$ returns 1, if one of the end points of $CS_i$ is close to an arbitrary end point of $CS_j$; otherwise it returns 0. It is given by:

$$D(CS_i, CS_j) = \begin{cases} \mathbf{1} & d(CS_i, CS_j) < th \\ \mathbf{0} & otherwise \end{cases} \tag{4}$$

In Equation (4), $d(CS_i, CS_j)$ is minimum distance between two end points $CS_i$ and $CS_j$ and ''$th$'' is the error tolerance.

Another measure $\Theta(CS_i, CS_j)$ is computed from $D(CS_i, CS_j)$ which shows if the gradients of tangents of the two segments are close enough at the end points. This measure is derived on the basis that there is no big difference in gradients of tangents of neighbor points on the same ellipse contour. The term $\Theta(CS_i, CS_j)$ is defined as:

$$\Theta(CS_i, CS_j) = \frac{1}{1 + \frac{|\theta_i - \theta_j|}{c}} \tag{5}$$

In Equation (5), $\theta_i$ and $\theta_j$ are gradients of tangents at the end points $CS_i$ and $CS_j$ that lie sufficiently close to each other, $c$ is a constant value for normalization ($\pi/\mathbf{2}$, it is maximum difference of gradients and 1 in the denominator prevents division by 0).

The merging measurement for the curve segments is determined using merging measurement ($MM$) as follows:

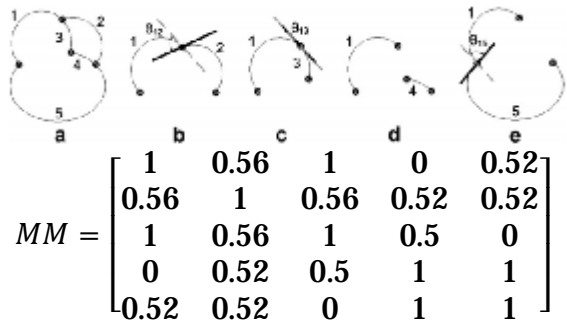

$$MM = \begin{bmatrix} 1 & 0.56 & 1 & 0 & 0.52 \\ 0.56 & 1 & 0.56 & 0.52 & 0.52 \\ 1 & 0.56 & 1 & 0.5 & 0 \\ 0 & 0.52 & 0.5 & 1 & 1 \\ 0.52 & 0.52 & 0 & 1 & 1 \end{bmatrix}$$

### C. Counting of size of Red Blood Cell (RBC)

By counting all the pixels of any selected RBC we can get the number of pixels that are in one micron dividing by diameter of RBC which is 6-9 microns. For this study RBC diameter is 8 microns.

$$1\ mm = 3.77\ pixels$$
$$1\ micron = 0.001\ mm$$
digital microscope with magnification of 1000x
$$1\ micron = 0.001x1000 = 1\ mm$$
$$\text{diameter RBC} \cong 8\ micron = 8\ mm$$
$$\cong 8 \times 3.77$$
$$\cong 30\ pixels$$

*D. Feature Extraction*

The diagnosis of acute leukemia became apparent after the patient evaluation and examination of the blood smear. Blast cells or immature cells usually exist in the majority of patients with AML, but can occur in some patients, no blast cells can be found. The presence of Auer rods in the blast can help in the diagnosis of AML. Some Auer rods can be found in some of the early cells of AML M3. According to Dr. Ninie Yusida Sp.PK.MSc, Chairman of Pathology Laboratory in Karang Anyar General Hospital, Auer is a form of granules which can be found in the promyelocyte cell types of the disease in the early diagnosis of AML M3. Examples of myeloblasts and promyelocyte cell images can be shown in Figure 3 and sample images of blood cells are detected AML M3 can be shown in Figure 3, (Bell, A &Sallah, S , 2005)

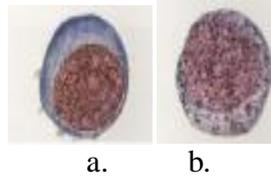

a.          b.

**Figure 3.** a.Myeloblast, b. Promyelocyte

Figure 4 shows morphological difference determined by FAB in Acute Lymphoblastic Leukemia cases (ALL), ( Labati, R. D, Piuri, V and Scotti, F, 2011)

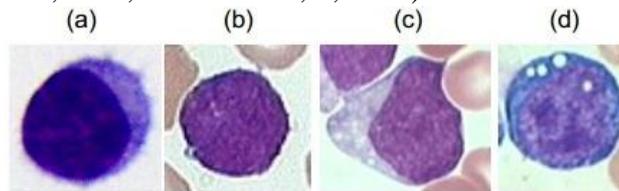

**Figure 4.** Morphological difference of blast cells. (a) Healthy lymphocyte, (b) L1, (c) L2, (d) L3.

Feature Extraction process is a process to obtain the characteristics or parameters that will be used to detect ALL, AML M3 or Healthy Cells. Characteristics to be searched is the WBC area, nuleus ratio and granule ratio. WBC area is the area of WBC, the nucleus ratio is the ratio between the area of the cell nucleus with the area of WBC. Feature extraction stage is as follows:

*a. Counting total count of pixels in WBC nucleus*
After detecting each WBC area then can be calculated area of each WBC nucleus. First we have to count number of pixel inside nucleus of WBCthen the area of WBC nucleus can be calculated. Area of WBC nucleus is calculated after calculation of RBC area.

*b. Counting Granule Ratio*
To obtain granule ratio, first will be calculated the number of pixels contained in granules. After that the number of pixels contained in granules will be compared with the number of pixels contained in the nucleus of WBC. In Figure 5 WBC granules are shown by red arrow.

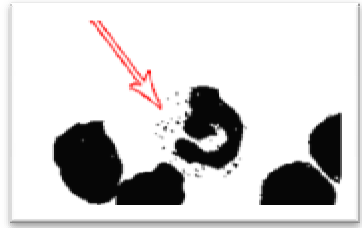

**Figure 5.** Granules are shown using red arrow.

*c. Area WBC and Nucleus WBC*

After counting total pixel count in WBC, total pixel count in WBC nucleus and the number of pixels in one micron. WBC area and nucleus areaare calculated by the following equation. For this study RBC diameter is taken as 8 microns.

$$RBC\ Radius\ = \frac{RBC\ Diameter}{2} = 4\ micron;$$ (6)

$$RBC\ Area\ = r^2 \times \pi = 4^2 \times 3.14 \cong 50\ micron^2;$$ (7)

$$Total\ pixels\ in\ 1\ micron^2 = \frac{Total\ pixels\ in\ 1\ RBC}{RBC\ Area};$$ (8)

$$Area\ WBC = \frac{Total\ pixels\ in\ 1\ WBC}{Total\ pixels\ in\ 1\ micron^2};$$ (9)

$$WBC\ Diameter\ = 2 \times \sqrt{\frac{WBC\ Area}{3.14}};$$ (10)

*D. Fuzzy Rule Based*

Fuzzy logic is a method to formalize the human capacity of imprecise reasoning. Such reasoning represents the human ability to reason approximately and judge under uncertainty. In fuzzy logic all truths are partial or approximate. In this sense, this reasoning has also been termed interpolative reasoning, where the process of interpolating between the binary extremes of true and false is represented by the ability of fuzzy logic to encapsulate partial truths (Ross, T J, 2010).

Fuzzy Inference System is also known as Fuzzy Rule Based that is used in this study is the zero-order Sugeno method. This method is used to create a rule or regulation of the input variables to the output variable. Input variable is the result of feature extraction, namely WBC Area, Nucleus Ratio, and Granule Ratio. While the output variable is the weighted average, percentage of sicknessof blood in image sample, so for further process it could be identified as ALL or AML, or Both Negative with a certain threshold of Weight Average.

The fuzzy level of understanding and describing a complex system is expressed in the form of a set of restrictions on the output based on certain conditions of the input. Restrictions are generally modeled by fuzzy sets and relations. These restriction statements are usually connected by linguistic connectives such as "and," "or," or "else" as shown in the example below

| Fuzzy Rule-Based System |
| --- |
| Rule1:   *IF x is $A^1$ and $A^2$ ... and $A^L$ THEN y is $B^s$* |
| Rule2:   *IF x is $A^1$ OR x is $A^2$ ... OR x is $A^L$ THEN y is $B^s$* |
| ⋮ |

> Ruler:    IF condition $C^r$, THEN restriction $R^r$

### a. Membership Function

In this research will be use a triangular curve representation. Triangular curve representation figure and equationsare as follows:

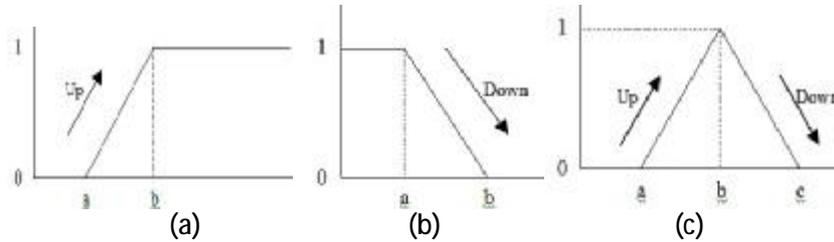

**Figure 6.** Triangular curve representation figures (a,b,c)

$$\mu_{Up}[x] = \begin{cases} 0; & x < a \\ \frac{x-a}{b-a}; & a \le x \le b \\ 1; & x > b \end{cases} \tag{11}$$

$$\mu_{Down}[x] = \begin{cases} 0; & x > b \\ \frac{b-x}{b-a}; & a \le x \le b \\ 1; & x < a \end{cases} \tag{12}$$

$$\mu_{Regular}[x] = \begin{cases} 0; & x < a \ or \ x > c \\ \frac{x-a}{b-a} & a \le x \le b \\ \frac{c-x}{c-b} & b \le x \le c \\ 1; & x = b \end{cases} \tag{13}$$

Equation 11, 12 and 13 is the triangular curve representation equation on the way up, down and regular (up and down).

### b. Sugeno Methode zero order

Sugeno method is computationally effective and works well with optimization and adaptive techniques, which makes it very attractive in control problems, particularly for dynamic nonlinear systems.MichioSugeno used a single spike, a singleton, as the membership function of the rule consequent. A singleton, or more precisely a fuzzy singleton, is a fuzzy set with a membership function that is unity at a single particular point on the universe of discourse and zero everywhere else. Zero - order Sugeno fuzzy model applies fuzzy rules in the following form:

IF x is A
AND y is B
THEN z is k

In this case, the output of each fuzzy rule is constant. All consequent membership functions are represented by singleton spikes.

III.     **Results and Analysis**

*A. Result of Feature Extraction and Fuzzy Membership*

The result of this research are :

a. *Membership of WBC Area*

Here is a table and membership function curve of the WBC area, can be shown in Table 1 and Figure 7.

TABLE I
MEMBERSHIP OF WBC AREA

| Diameter (micron) | Description |
|---|---|
| 6-10-15 | Small |
| 10-15-30 | Medium |
| 15-25-60 | Big |

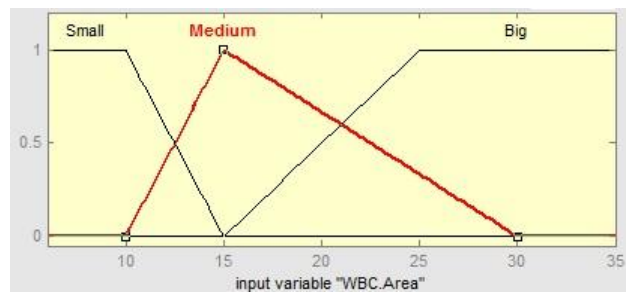

**Figure 7.** Membership function curve of WBC Area

*b. Membership of Nucleus Ratio*

Here is a table and membership function curve of the nucleus ratio, can be shown in Table 2 and Figure 8

TABLE II

MEMBERSHIP OF NUCLEUS RATIO

| Ratio | Description |
|---|---|
| 0.0- 0.2- 0.3 | Small |
| 0.2-0.5-0.7 | Medium |
| 0.6-0.75-1.0 | Big |

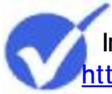

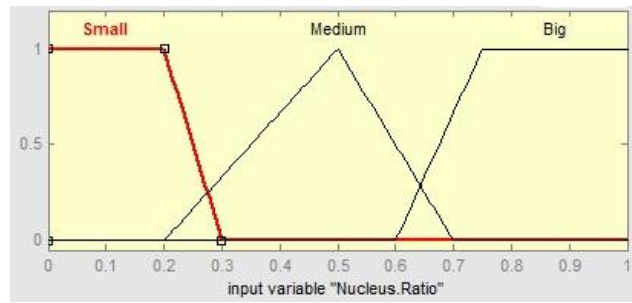

**Figure 8.** Membership function curve of Nucleus Ratio

*c. Membership of Granule Ratio*

Here is a table of membership of granule ratio and granule ratio membership functions curve, can be shown in Table 3 and Figure 9:

TABLE III

MEMBERSHIP OF GRANULA RATIO

| Granula Ratio | Description |
|---------------|-------------|
| 0-0.1-0.2 | Small |
| 0.1-0.3-1 | Big |

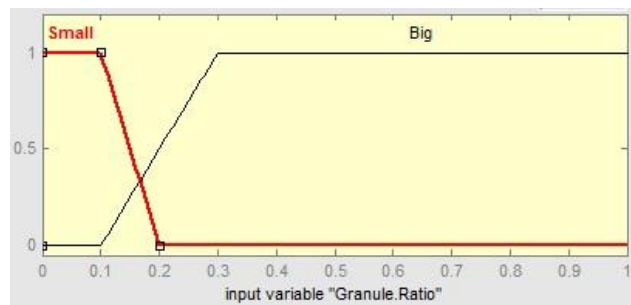

**Figure 8.** Membership function curve of Granula Ratio.

*d. Fuzzy Rule*

The next thing to be done is to create an input rule and output results. Create rule or rules as shown in Table 4. Rules in Table 4 consists of three inputs and one output variable 0, 1, and 2 which shows if output is 0 then it is not detected as ALL neither AML M3, if the output results 1 then it is detected as ALL and the last one if it is 2 then it is detected as AML M3.

TABLE IV

SCHEME OF FUZZY RULE BASED INPUT AND OUTPUT WITH SUGENO
METHOD

| No | WBC Area | Nucleus Ratio | Granule Ratio | Result |
|---|---|---|---|---|
| 1 | Small | - | Small | 0 |
| 2 | Medium | Small | Small | 0 |
| 3 | Medium | Small | Big | 2 |
| 4 | Medium | Medium | Small | 0 |
| 5 | Medium | Medium | Big | 2 |
| 6 | Medium | Big | Small | 1 |
| 7 | Big | Small | Small | 0 |
| 8 | Big | Small | Big | 2 |
| 9 | Big | Medium | Small | 0 |
| 10 | Big | Medium | Big | 2 |

The test results of 29 images of blood cells are known as Positive ALL only1 image that could not be identified by the system, from 50 images of blood cells known as AML M3 Positive, there are 6 images that areidentified incorrectly by the system, and out of 25 images that are ALL negative and AML negative, there are 5 images that were identified incorrectly by the system, and there are 5 images that could not be identified by the system. WA values are used to determine an image:The WA value that identifies image as ALL negative and AML negative (None) is $0 \leq WA < 0.5$, to identify ALL it is $0.5 \leq WA \leq 1$, while for identifying AML is $1 \leq WA \leq 2$, the accuracy of the system calculated as follows:

$$.\textbf{Accuracy} = \left(1 - \frac{(TW+TU)}{TI}\right) \times 100\% \tag{14}$$

Explanation :

TW = Total Images that are identified as wrongly.

TU = Total Unidentified Images.

TI = Total Images that used for test.

From 104 images obtained test results show that 87 images Detected True, 11 images Detected Falseand 6 images could not be detected , and the system accuracy is as follows:

$$\textbf{Accuracy} = \left(1 - \frac{(11 + 6)}{104}\right) \times 100\% = \textbf{83,65}\%$$

*B. Analysis*

Analysis of of image that is detected wrong is as follows:

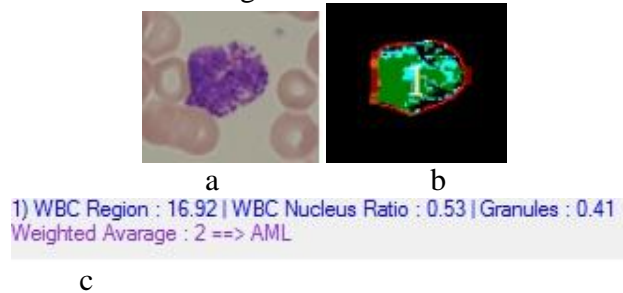

**Figure 9.** Example of incorrectly detected result, image Im040_0.jpg (Healthy) (a) The original image of lymphocyte,(b) WBC detection result, (c) result of morphological characteristics and Fuzzy Rule.

Following is a discussion of the example shown in Figure 9: As in the membership function of WBC Area, Nucleus Ratioand Granule Ratio for image Im040.0.jpg WBC Area = 16.92, Nucleus Ratio = 0.53 and Granule Ratio= 0.41 also can be said that the WBC Area = Medium or Large , Nucleus Ratio = Medium, Granule Ratio = Large. With these criteria there are only two possible rules as follows:

1. IF WBC area = Medium && Nucleus Ratio = Medium && Granule Ratio = Big THEN 2
2. IF WBC area = Big && Nucleus Ratio = Medium && Granule Ratio = Big THEN 2

Solution for Rule 1:

Membership functions for WBCArea - Medium (16.92) with a triangular representation curve down has a limit of $10 \leq x \leq 30$

$$\frac{c - x}{c - b} = \frac{30 - 16.92}{30 - 10} = \frac{13.08}{20} = 0.654;$$

Membership functions for Nucleus Ration - Medium (0.53) with a triangular representation curve down has a limit of $0.2 \leq x \leq 0.7$

$$\frac{c - x}{c - b} = \frac{0.7 - 0.53}{0.7 - 0.2} = \frac{0.17}{0.5} = 0.34;$$

Membership functions for Granule Ration - Big (0.41) with a triangular representation curve down has a limit of $0.1 \leq x \leq 0.2$

$$x \geq 0.2 \ the\ membership\ value\ is\ \ 1;$$

Berikut dapat dihitung untuk *rule* 2:

Membership functions for WBC Area - Big (16.92) with a triangular representation curve down has a limit of $15 \leq x \leq 60$

$$\frac{x - a}{b - a} = \frac{16.92 - 15}{60 - 15} = \frac{1.92}{45} = 0.04267;$$

Membership functions for Nucleus Ration - Medium (0.53) with a triangular representation curve down has a limit of $0.2 \leq x \leq 0.7$

$$\frac{c - x}{c - b} = \frac{0.7 - 0.53}{0.7 - 0.2} = \frac{0.17}{0.5} = 0.34;$$

Membership functions for Granule Ratio - Big (0.41) with a triangular representation curve down has a limit of $0.1 \leq x \leq 0.2$

$$x \geq 0.2 \ \ membership\ value\ is\ 1;$$

Calculated Waighted Avarage with Rule1 (min) and Rule 2 (min) is:

$$WA = \frac{(2*0.34) + (2*0.04267)}{0.34 + 0.04267} = \frac{0.68 + 0.08534}{0.38267} = 2;$$

So the result of Rule 1 and Rule 2 is WA = 2. Obraines result shows that digital images im040.0.jpg (healthy image/ Healthy) is detected as AML by the system.

## IV.    Conclusions

This study has shown that is able to identify leukemia disease especially Acute leukemia ALL and AML type M3 through morphological characteristicsusing image processing. Morphological features used in this study are WBC Area, Nucleus Ratio and Granule Ratio. Accuracy of the results of testing reached 83, 65% of 104 digital image samples of blood used for testing.